\documentclass{article}
\usepackage{spconf,amsmath,graphicx}

\usepackage{graphicx}
\usepackage{amsmath}
\usepackage{color}
\usepackage{booktabs}
\usepackage{wrapfig}
\usepackage{amsfonts}

\newcommand{\ie}{\emph{i.e. }}
\newcommand{\etc}{\emph{etc}}

\title{Balanced Binary Neural Networks with Gated Residual}
\name{Mingzhu Shen\textsuperscript{1}, Xianglong Liu\textsuperscript{1,}\thanks{Corresponding author}, Ruihao Gong\textsuperscript{1}, Kai Han\textsuperscript{2}}
\address{\textsuperscript{1} State Key Laboratory of Software Development Environment, Beihang University, China\\
	\textsuperscript{2} State Key Lab of Computer Science, Institude of Software, CAS; UCAS \\
	shenmingzhu@buaa.edu.cn\\
	\{xlliu, gongruihao\}@nlsde.buaa.edu.cn \\
	hankai@ios.ac.cn}
\begin{document}
%
\maketitle
\begin{abstract}
	Binary neural networks have attracted numerous attention in recent years. However, mainly due to the information loss stemming from the biased binarization, how to preserve the accuracy of networks still remains a critical issue. In this paper, we attempt to maintain the information propagated in the forward process and propose a \textbf{B}alanced \textbf{B}inary Neural Networks with \textbf{G}ated Residual (BBG for short). First, a weight balanced binarization is introduced and thus the informative binary weights can capture more information contained in the activations. Second, for binary activations, a gated residual is further appended to compensate their information loss during the forward process, with a slight overhead. Both techniques can be wrapped as a generic network module that supports various network architectures for different tasks including classification and detection. The experimental results show that BBG-Net performs remarkably well across various network architectures such as VGG, ResNet and SSD with the superior performance over state-of-the-art methods.
\end{abstract}
\section{Introduction}
Deep neural networks (DNNs), especially deep convolution neural networks (CNNs), have been well demonstrated in a wide variety of computer vision applications. However, most of these advanced deep CNN models requires expensive storage and computing resources, and cannot be easily deployed on portable devices such as mobile phones, cameras, \etc.

In recent years, a number of approaches have been proposed to learn portable deep neural networks, including multi-bit quantization~\cite{google8bit}, pruning~\cite{wang2017towards}, and lightweight architecture design~\cite{zhang2018shufflenet}, knowledge distillation~\cite{zagoruyko2016paying}. Among them, quantization based methods~\cite{google8bit,zhang2018lq} represent the weights and activations in the network with a very low precision, and thus can yield highly more compact DNN models compared to those floating-point counterparts. In an extreme case where both the weights and activations are quantized into one-bit values, the conventional convolution operations can be efficiently achieved via bitwise operations~\cite{rastegari2016xnor}. Ihe resulting decrease in the storage and acceleration in the inference are therefore appealing to the community. Since the proposition of binary neural network, many works have been done to address the performance drop and improve expression ability of binarized networks. Bireal-Net~\cite{bi-real} proposes using additional shortcut and different approximation of sign function in the backward pass. PCNN~\cite{gu2019projection} employs a projection matrix to help with the network training. CircConv~\cite{liu2019circulant} rotate binary weight by three times and calculate feature map with four binary weights and merge them together. BENN~\cite{zhu2019binary} ensembles multiple standard binary networks to improve performance. AutoBNN~\cite{shen2019searching} employs genetic algorithm to search for binary neural network architectures. Most of these methods have complicated training pipeline and even increase the FLOPs to achieve better results.
\begin{figure}
	\centering
	\small
	\includegraphics[width=1.0\linewidth]{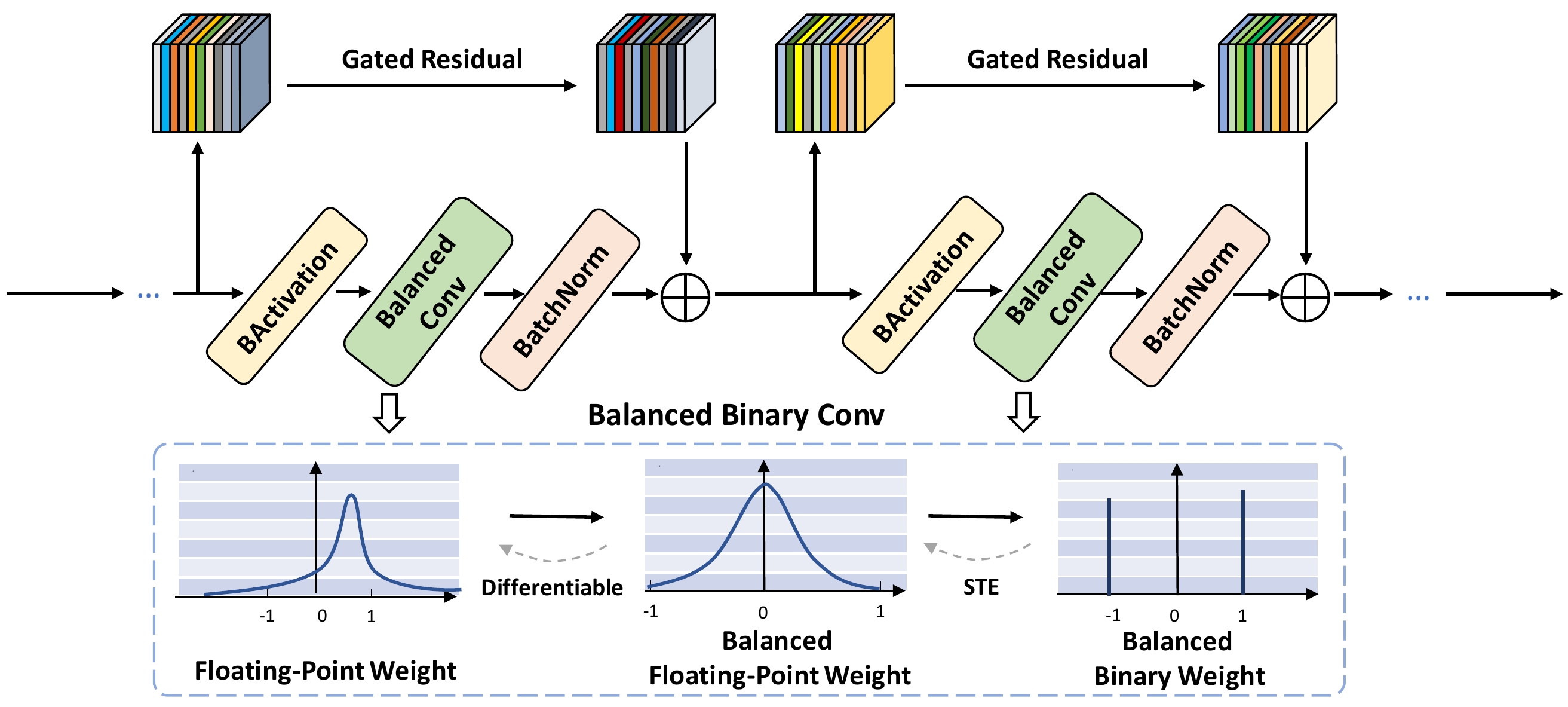}
	\vspace{-0.2in}
	\caption{The diagram of the proposed method for learning binary neural networks by exploiting balanced weights binarization and gated residual to reconstruct information loss.}
	\label{1}
	\vspace{-0.2in}
\end{figure}

Although much progress has been made on binarizing DNNs, the existing quantization methods still cause a significantly large accuracy drops, compared with the full-precision models. This observation indicates that the information propagates through the full-precision network are largely lost, when using the binary representation. 
To address this problem and maximally preserve the information, in this paper we propose the \textbf{B}alanced \textbf{B}inary Network with \textbf{G}ated Residual (BBG for short), which learns the balanced binary weights, and reduce binarization loss for activations by a gated residual path. Fig.~\ref{1} demonstrates the whole structure of our BBG. We re-parameterize the weight and devise a linear transformation to replace the standard one, which can be easily implemented and learnt. To compensate the information loss when binarizing activations, the gated residual path employs a lightweight operation to use the channel attention information of floating-point activations to further reconstruct the information flow across layers. We evaluate our BBG method on the image classification and object detection benchmarks, and the experimental results show that it performs remarkably well across various network architectures, outperforming state-of-the-art.

\section{The Proposed Method}

\subsection{Maximizing Entropy with Balanced Binary Weights}
In Binary Neural Networks, the most important training parameter is the non-differentiable discrete binary weights. This discrete property of binary weights brings troublesome problems to the training of the network. To preserve the information of binary weights, we propose to maximize entropy of binary weight in the training process. Directly optimize the above regularization is hard. Instead, we approximate the optimal solution for it by making the expectation of binary weights to be zero to, \ie ${\mathbf{w^b}}^\top\mathbf{1}=0$. 

Hence, we first center the real-valued weights and then quantize them into binary codes. More specifically, we use a proxy parameter $\mathbf v\in\mathbb R^{d \times d}$ to get $\mathbf w$ and then quantize it into binary codes ${\mathbf{w^b}}$. In the convolutional layer, we express the weight $\mathbf w$ in terms of the proxy parameter $\mathbf v$ using:	\begin{equation}\label{eq:bwq}
\mathbf w = \mathbf v-\frac{1}{d}\mathbf 1(\mathbf 1^\top\mathbf v),
\end{equation}
After balanced weight normalization, we can perform binary quantization on the floating-point weight $\mathbf w$. Subsequently, the forward pass and backward pass of binary weights are as follows:
\begin{equation}\label{binarize}
\begin{aligned}
\text{Forward:} &\mathbf{w^b} = \text{sign}(\mathbf w) \times E(|\mathbf w|),\\
\text{Backward:} & \frac{\partial{\mathcal{L}}}{\partial{\mathbf {w^b}}} = \frac{\partial{\mathcal{L}}}{\partial{\mathbf w}},
\end{aligned}
\end{equation}
where $\text{sign}(\cdot)$ is sign function that outputs $+1$ for positive numbers and $-1$ otherwise and $E(|\cdot|)$ calculates the mean of absolute value.

In our balanced weight quantization, the parameter updating is completed based on the proxy parameters $\mathbf v$, and thus the gradient signal can back-propagate through the normalization process. In the whole process, $\mathbf v$ and $\mathbf w$ are floating-point, and we update $\mathbf v$ on training process and only ${\mathbf{w^b}}$ is needed on inference. 

\subsection{Reconstructing Information Flow with Gated Residual}	
In the binarization of activation, we first clip the value range of activations $\mathbf x$ into $[0,1]$ and use a round function to binarize activations. Therefore, the forward pass and backward pass for binary activations are as follows:
\begin{equation}
\label{bpactivation}
\begin{aligned}
\text{Forward:} &\mathbf{x^b} = \text{round}(\text{clip}(\mathbf x,0,1)),\\
\text{Backward:} & \frac{\partial{\mathcal{L}}}{\partial{\mathbf {x^b}}} = \frac{\partial{\mathcal{L}}}{\partial{\mathbf x}}\mathbb{I}_{0< \mathbf x<1},
\end{aligned}
\end{equation}
where $\mathbb{I}_{0< \mathbf x<1}$ means if elements of $\mathbf x$ is in the range of $[0,1]$, then it is 1, otherwise 0.

Binarizing activations results in much larger loss of precision than binary weights. All activation values of different channels are quantized to 0 or 1, without considering the differences among the channels. The quantization error caused by binary layers is accumulated layer by layer. To address this problem, we further propose a new module named gated residual to reconstruct information flow in channel-wise manner, during the forward propagation. Our gated residual employs the floating-point activations to reduce quantization error and recalibrate features.
\begin{figure}
	\centering
	\small
	\includegraphics[width=1\linewidth]{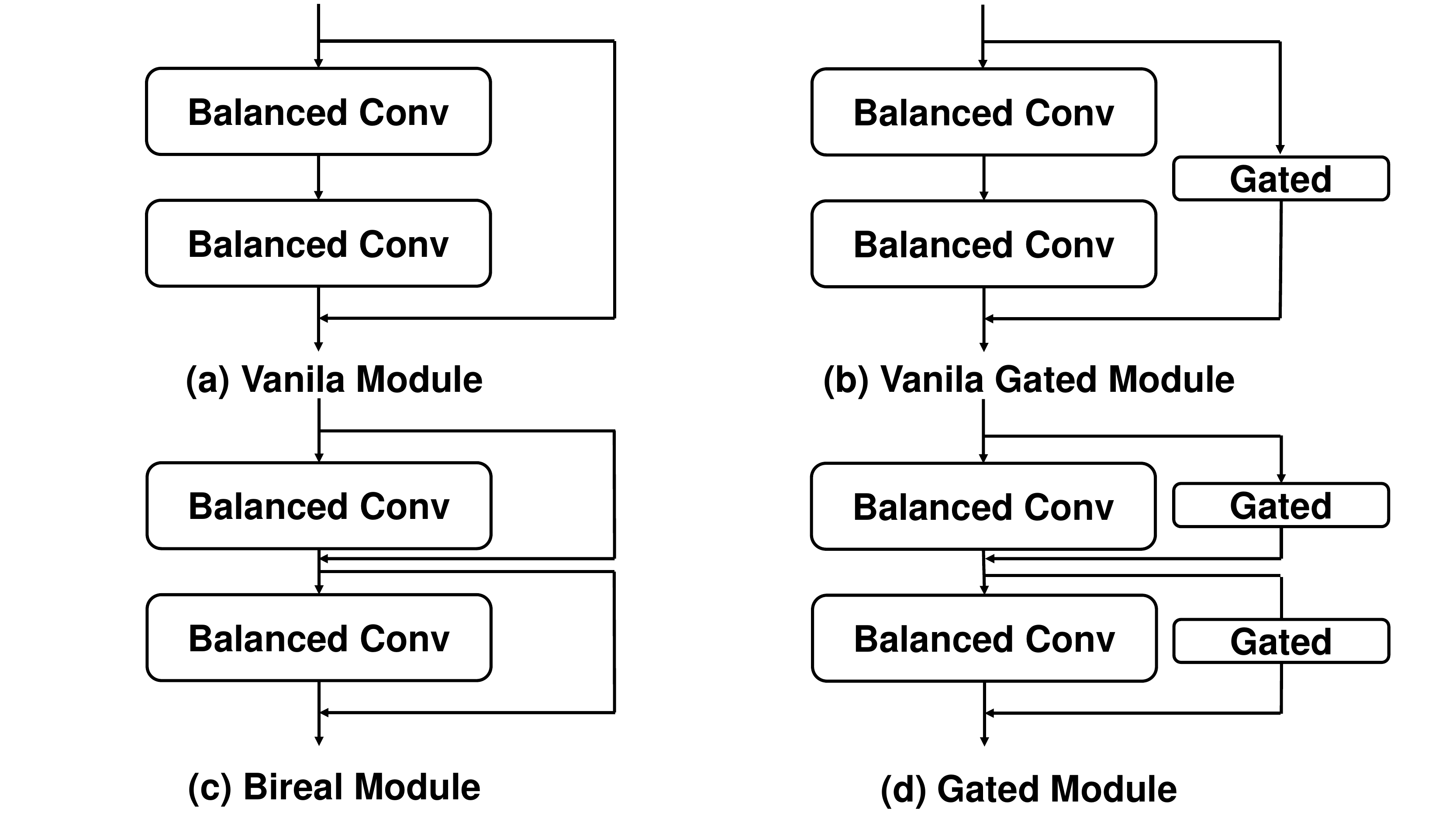}
	\vspace{-0.2in}
	\caption{The four different kinds of modules are utilized to construct ResNet networks.}
	\label{module}
	\vspace{-0.2in}
\end{figure}

Subsequently, we propose a new designed layer with gated weights $\mathbf s=[s_1,...,s_c]\in \mathbb R^{c}$ that learn the channel attention information of floating-point input feature map $\mathbf x = [\mathbf {x_1},..., \mathbf{x_c}] \in \mathbb R^{c\times h \times w}$ in a binary convolution layer ($c$, $h$, $w$ means channels, height and width respectively). The operation on the $i$th channel of the input feature map $\mathbf {x_i}$ is defined as follows:	
\begin{equation}\label{scale}
r(\mathbf {x_i},s_i)= s_i\mathbf{x_i}
\end{equation}
Based on the gated residual, the output feature map $\mathbf y \in \mathbb R^{c\times h \times w}$ can be recalibrated, enhancing the representation power of the activations. The operation in the gated module can be written in the following form:
\begin{equation}\label{highway}
\mathbf y = \mathcal F(\mathbf x) + r(\mathbf x,\mathbf s)
\end{equation}	
where $\mathcal F(x)$ means the operation in the main path including activation binarization, balanced convolution and BatchNorm in total.

With gated residual layer, the overall structure of gated module is shown in the Figure~\ref{module}(d). We initialize weight of gated residual with $1$ which is the same with identity shortcut of vanilla ResNet. In training process, the gated residual learns to distinguish the channels and eliminate the less useful channels. Beside reconstructing the information flow in the forward process, this path also acts as an auxiliary gradient path for activations in the backward propagation. Usually, the STE used in the backward pass to approximate discrete binary functions results in severe gradient mismatch problem. Fortunately, with learnable gated weights, our gated residual can also reconstruct gradient in the following way:

\begin{equation}
\frac{\partial{\mathbf y}}{\partial{\mathbf x}} = \frac{\partial{\mathcal F(\mathbf x)}}{\partial{\mathbf x}}+\mathbf s
\end{equation}

In terms of computational complexity and memory limitation in the nature of the binary networks, the additional operations for designing a new module need to be as small as possible. The structure of HighwayNet~\cite{srivastava2015highway} requires a full-precision weights that is as large as the weight in the convolutional layer, which is unacceptable. Compared to SENet~\cite{hu2018squeeze}, SE module is a correction to the output after the convolution layer, and we argue that it is better to make use of the unquantified information. Similarly, our FLOPs is only $c \times w \times h$ and is much smaller than the SE module. When the number of channels increases and the reduction decreases, the amount of FLOPs required by the SE module will be much more. 	

\section{Experiments}
In this section, to verify the effectiveness of our proposed BBG-Net, we conduct experiments on both image classification and object detection task.
\subsection{Datasets and Implementation Details}

\textbf{Datasets.}
In our experiments, we adopt three common image classification datasets: CIFAR-10/100, and ILSVRC-2012 ImageNet. We also evaluate our proposed method on the object detection task with a standard detection benchmark, Pascal VOC datasets named VOC2007, and VOC2012. 

\textbf{Network Architectures and Setup.}We conduct our experiments on popular and powerful network architectures ResNet~\cite{he2016deep} and Single Shot Detector(SSD\cite{liu2016ssd}). For fair comparison with the existing methods, we respectively choose  ResNet-20, ResNet-18 as the baseline model on CIFAR-10/100 and ImageNet datasets. And we verify SSD with different backbones, i.e. VGG-16~\cite{simonyan2014very} and ResNet-34 in Pascal VOC datasets. As for hyper-parameter, we mostly follow the same setup in the original papers and all our models are trained from scratch. Note that we do not quantize the first and last layer and we also do not quantize down-sample layers as suggested by many previous work~\cite{bethge2019back}. And we only quantize backbone network in SSD.
\begin{table}
	\centering
	\small
	\renewcommand\arraystretch{1.05}
	\caption{Performance comparison of our proposed methods with vanilla binary networks on ResNet-20 networks on CIFAR-10 validation set.}
	\label{Balanced Results Comparison}
	\vspace{0.3em}
	\begin{tabular}{|c||c|c|c|}
		\hline
		W/A&Weight&Residual&Acc.(\%)\\
		\hline
		\hline
		32/32&FP&FP&92.1\\
		\hline
		1/1&Vanilla&Identity&84.13\\
		1/1&Balanced&Identity&84.71\\
		1/1&Vanilla&Gated&84.89\\
		1/1&Balanced&Gated&\textbf{85.34}\\
		\hline
	\end{tabular}
	\vspace{-0.2in}
\end{table}	 	

\subsection{Ablation Study}
Now we study how our balanced weight quantization and gated residual module affects the network's performance. In Table~\ref{Balanced Results Comparison}, we report the results of ResNet-20 on CIFAR-10, with and without balanced quantization or gated residual. In the performance comparison from the first two rows, the network with balanced quantization can obtain 0.6\% accuracy than that without this operation. From the whole table, we can easily observe that, balanced weights or gated residual brings accuracy improvement and together they work even better with 1.2\% accuracy improvement. It reveals that our proposed method faithfully helps pursue a highly accurate binary network.

\begin{table}
	\centering
	\small
	\renewcommand\arraystretch{1.05}
	\caption{Performance comparison of 4 different modules on ResNet-20 models on CIFAR-10/100 validation set.}
	\label{resnet-20 results}
	\vspace{0.5em}
	\begin{tabular}{|c||c|c|c|}
		\hline
		Method&Kernel Stage&CIFAR-10&CIFAR-100\\
		\hline
		\hline
		FP&16-32-64&92.1&68.1\\
		\hline
		Vanilla&16-32-64&84.71&53.37\\
		Vanilla Gated&16-32-64&84.96&55.24\\
		Bireal&16-32-64&\textbf{85.54}&55.07\\
		Gated&16-32-64&85.34&\textbf{55.62}\\
		\hline
		Vanilla&32-64-128&90.22&65.06\\
		Vanilla Gated&32-64-128&\textbf{90.71}&66.15\\
		Bireal&32-64-128&90.27&65.6\\
		Gated&32-64-128&90.68&\textbf{66.47}\\
		\hline 
		Vanilla&48-96-192&92.01&68.66\\
		Vanilla Gated&48-96-192&92.31&69.11\\
		Bireal&48-96-192&91.78&68.5\\
		Gated&48-96-192&\textbf{92.46}&\textbf{69.38}\\
		\hline
	\end{tabular}
	\vspace{-0.3in}
\end{table}

\subsection{Comparison with the State-of-the-Art}
\textbf{CIFAR-10/100 Dataset.}
In ResNet-20 on CIFAR-10/100 datasets, we further compare four different kinds of modules in Fig.~\ref{module}. As shown in Table~\ref{resnet-20 results}, in CIFAR-10 datasets, the accuracy improvement caused by Gated or module is less than 1\% while in a more challenging CIFAR-100 dataset, it adds up to over 2\%. Through the whole experiments, we can conclude that Vanilla Gated and Gated show superiority over Bireal and Vanilla modules. Especially when the network grows wider, Bireal Module even performs worse than Vanilla while Vanilla Gated and Gated consistently performs better. With the kernel stage of 48-96-192, our binary network matches the accuracy of full-precision networks in both CIFAR-10 and CIFAR-100 datasets.

\begin{figure}
	\centering
	\includegraphics[width=1\linewidth]{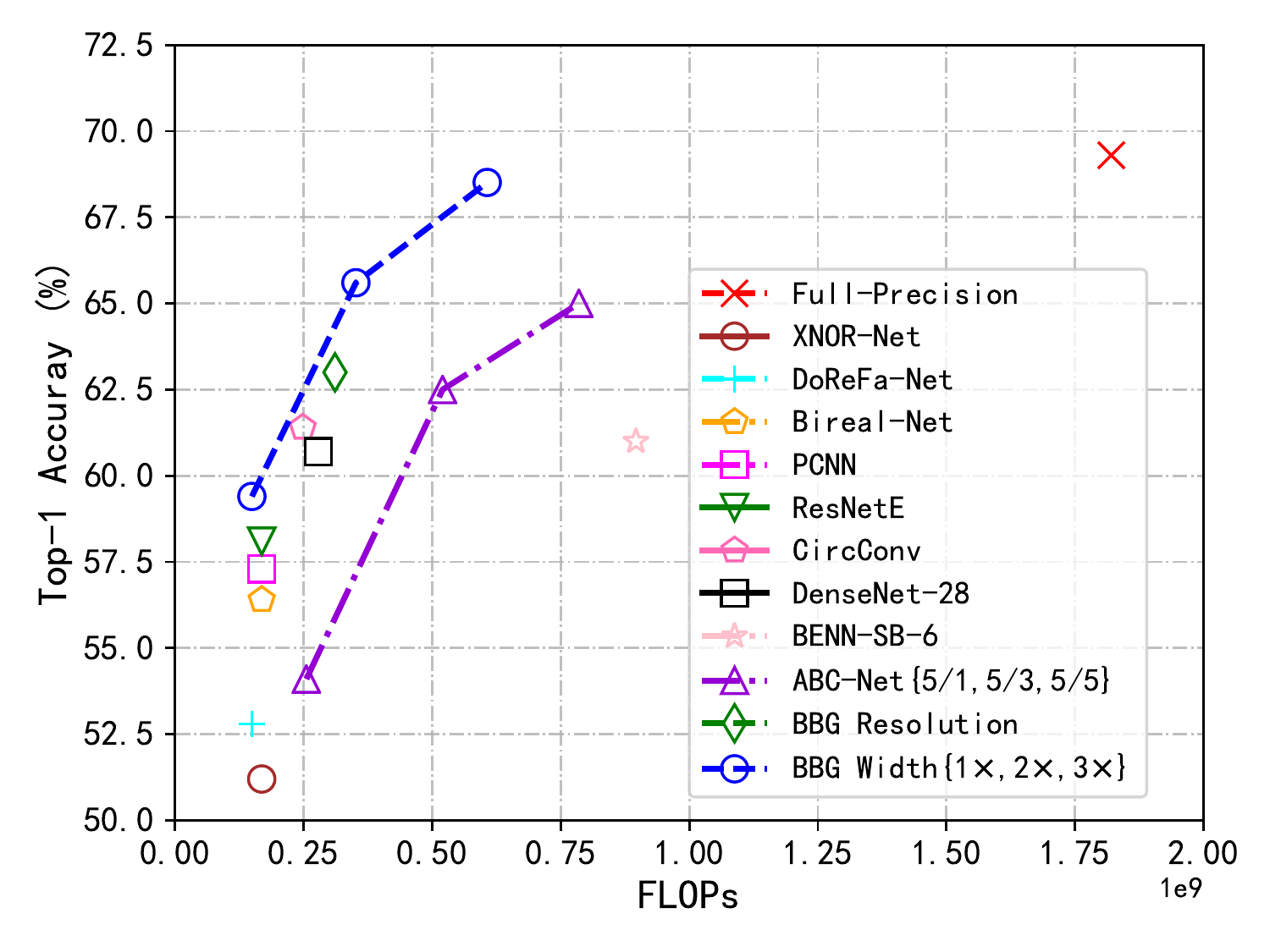}
	\vspace{-0.3in}
	\caption{Performance comparison with state-of-the-art methods and our network width and resolution exploration experiments.}
	\label{vis}
	\vspace{-0.2in}
\end{figure}

\textbf{ResNet-18 on ImageNet Dataset.}In Fig.~\ref{vis}, we compare our method with many binarization methods of recent years including XNOR-Net~\cite{rastegari2016xnor}, DoReFa-Net~\cite{zhou2016dorefa}, Bireal-Net~\cite{bi-real}, PCNN~\cite{gu2019projection}, and it further reveals stability of our proposed BBG-Net on larger datasets. Our method significantly outperforms all the other methods, with 1.3\% performance gain over the state-of-the-art ResNetE~\cite{bethge2019back}. 

We also improve the accuracy of binary networks by exploring network width and resolution in a simple but effective way. In the calculation of FLOPs, the binary layer is divided by $64$ following Bireal-Net\cite{bi-real}. In the network width experiments, we expand all channels of the original ResNet-18 by $2\times$ and $3\times$. Compared with the full-precision networks, our $3\times$ width models can achieve almost the same accuracy. We also compare with ABC-Net where $5/3$ means 5 binary bases for weight and 3 bases for activations, our methods consistently perform better than ABC-Net by a large margin. In resolution exploration, we employ a simple strategy that we remove the max pooling layer after the first convolution, which makes all hidden layers have $2\times$ feature maps compared with the original ones. Compared to CircConv~\cite{liu2019circulant} which employs four times binary weights, we have better results with 1.6\% accuracy improvement, while we have slightly higher FLOPs but the same memory. Our resolution accuracy is 2.3\% higher while the FLOPs is only $1.1\times$ of the FLOPs of DenseNet-28~\cite{bethge2019back}. BENN~\cite{zhu2019binary} ensembles 6 standard binary ResNet-18 which has nearly three times FLOPs of our resolution while the accuracy declines by 2\%.	

\begin{table}[h]
	\vspace{-0.2in}
	\centering
	\small
	\caption{Performance of binarized SSD on Pascal VOC dataset.}
	\vspace{0.3em}
	\renewcommand\arraystretch{1.0}
	\begin{tabular}{|c||c|c|c|c|}
		\hline
		Method&Backbone&Resolution&FLOPs&mAP(\%)\\
		\hline\hline
		FP&VGG-16&300&29986&74.3\\
		FP&ResNet-34&300&6850&75.5\\
		\hline
		XNOR&ResNet-34&300&362&55.1\\
		TBN&ResNet-34&300&464&59.5\\	
		BDN&DenseNet-37&512&1530&66.4\\
		BDN&DenseNet-45&512&1960&68.2\\
		\hline
		BBG&ResNet-34&300&362&\textbf{62.8}\\
		BBG&VGG-16&300&1062&\textbf{68.5}\\
		\hline
	\end{tabular}
	\label{ssd}
	\vspace{-0.2in}
\end{table}

\textbf{SSD on Pascal VOC Dataset.}
In object detection task, we compare our method with XNOR-Net~\cite{rastegari2016xnor}, TBN~\cite{wan2018tbn}, and BDN~\cite{bethge2019back} which includes DenseNet37 and DenseNet-45.In the comparison of ResNet-34 as its backbone, we outperform XNOR and TBN by 6.9\% and 2.5\%. It proves that our solution using 1-bit can preserve the stronger feature representation and maintain the better generalization ability than ternary neural networks. As for VGG-16, we only need half of FLOPs of binary DenseNet-45 to achieve slightly higher results and our model outperforms DenseNet-37 by 2\% with one thirds fewer FLOPs. The accuracy of our VGG-16 is only 5.8\% less than full-precision counterparts. The significant performance gain further demonstrates that our method can better preserve the information propagated in the network and help extract the most discriminative features for detection task.

\subsection{Deploying Efficiency}
Finally, we implement our method on mobile devices using a framework named daBNN~\cite{zhang2019dabnn}. The mobile device we use is Rasberry Pi 3B, which has a 1.2GHz 64-bit quad-core ARM Cortex-A53. As shown in Table~\ref{efficiency}, we implement DoReFa which binarizes downsample layers while we do not, the difference in inference time is only $2\text{ms}$ which can be ignored. Our proposed method can run $5.8\times$ faster than full-precision counterparts.

\begin{table}[h]
	\vspace{-0.2in}
	\centering
	\small
	\caption{Comparison of inference time of full-precision ResNet-18 and binary ResNet-18.}
	\vspace{0.3em}
	\renewcommand\arraystretch{1.0}
	\begin{tabular}{|c||c|c|c|}
		\hline
		Model&Full-Precision&DoReFa&BBG\\
		\hline
		\hline
		time(ms)&1457&249&251\\
		\hline
	\end{tabular}
	\vspace{-0.35in}
	\label{efficiency}
\end{table}
\section{Conclusion}
Binarization methods for establishing portable neural networks are urgently required so that these networks with massive parameters and complex architectures can be launched efficiently. In this work, we proposed a novel Balanced Binary Neural Network with Gated Residual, namely BBG-Net. 
Experiments conducted on benchmark datasets and architectures demonstrate the effectiveness of the proposed balanced weight quantization and gated residual for learning binary neural networks with higher performance but lower memory and computation consumption than the state-of-the-art methods.
\bibliographystyle{IEEEbib}
\bibliography{Template}

\begin{thebibliography}{10}

\bibitem{google8bit}
Jacob Benoit, Kligys Skirmantas, Chen Bo, Zhu Menglong, Tang Matthew, Howard
  Andrew, Adam Hartwig, and Kalenichenko Dmitry,
\newblock ``Quantization and training of neural networks for efficient
  integer-arithmetic-only inference,''
\newblock in {\em CVPR}, 2018, pp. 2704--2713.

\bibitem{wang2017towards}
Yunhe Wang, Chang Xu, Jiayan Qiu, Chao Xu, and Dacheng Tao,
\newblock ``Towards evolutional compression,''
\newblock {\em arXiv preprint arXiv:1707.08005}, 2017.

\bibitem{zhang2018shufflenet}
Xiangyu Zhang, Xinyu Zhou, Mengxiao Lin, and Jian Sun,
\newblock ``Shufflenet: An extremely efficient convolutional neural network for
  mobile devices,''
\newblock in {\em Proceedings of the IEEE Conference on Computer Vision and
  Pattern Recognition}, 2018, pp. 6848--6856.

\bibitem{zagoruyko2016paying}
Sergey Zagoruyko and Nikos Komodakis,
\newblock ``Paying more attention to attention: Improving the performance of
  convolutional neural networks via attention transfer,''
\newblock {\em arXiv preprint arXiv:1612.03928}, 2016.

\bibitem{zhang2018lq}
Dongqing Zhang, Jiaolong Yang, Dongqiangzi Ye, and Gang Hua,
\newblock ``Lq-nets: Learned quantization for highly accurate and compact deep
  neural networks,''
\newblock in {\em Proceedings of the European Conference on Computer Vision
  (ECCV)}, 2018, pp. 365--382.

\bibitem{rastegari2016xnor}
Mohammad Rastegari, Vicente Ordonez, Joseph Redmon, and Ali Farhadi,
\newblock ``Xnor-net: Imagenet classification using binary convolutional neural
  networks,''
\newblock in {\em European Conference on Computer Vision}. Springer, 2016, pp.
  525--542.

\bibitem{bi-real}
Zechun Liu, Baoyuan Wu, Wenhan Luo, Xin Yang, Wei Liu, and Kwang-Ting Cheng,
\newblock ``Bi-real net: Enhancing the performance of 1-bit cnns with improved
  representational capability and advanced training algorithm,''
\newblock in {\em ECCV}, 2018.

\bibitem{gu2019projection}
Jiaxin Gu, Ce~Li, Baochang Zhang, Jungong Han, Xianbin Cao, Jianzhuang Liu, and
  David Doermann,
\newblock ``Projection convolutional neural networks for 1-bit cnns via
  discrete back propagation,''
\newblock in {\em Proceedings of the AAAI Conference on Artificial
  Intelligence}, 2019, vol.~33, pp. 8344--8351.

\bibitem{liu2019circulant}
Chunlei Liu, Wenrui Ding, Xin Xia, Baochang Zhang, Jiaxin Gu, Jianzhuang Liu,
  Rongrong Ji, and David Doermann,
\newblock ``Circulant binary convolutional networks: Enhancing the performance
  of 1-bit dcnns with circulant back propagation,''
\newblock in {\em Proceedings of the IEEE Conference on Computer Vision and
  Pattern Recognition}, 2019, pp. 2691--2699.

\bibitem{zhu2019binary}
Shilin Zhu, Xin Dong, and Hao Su,
\newblock ``Binary ensemble neural network: More bits per network or more
  networks per bit?,''
\newblock in {\em Proceedings of the IEEE Conference on Computer Vision and
  Pattern Recognition}, 2019, pp. 4923--4932.

\bibitem{shen2019searching}
Mingzhu Shen, Kai Han, Chunjing Xu, and Wang Yunhe,
\newblock ``Searching for accurate binary neural architectures,''
\newblock {\em arXiv preprint arXiv:1909.07378}, 2019.

\bibitem{srivastava2015highway}
Rupesh~Kumar Srivastava, Klaus Greff, and J{\"u}rgen Schmidhuber,
\newblock ``Highway networks,''
\newblock {\em arXiv preprint arXiv:1505.00387}, 2015.

\bibitem{hu2018squeeze}
Jie Hu, Li~Shen, and Gang Sun,
\newblock ``Squeeze-and-excitation networks,''
\newblock in {\em Proceedings of the IEEE conference on computer vision and
  pattern recognition}, 2018, pp. 7132--7141.

\bibitem{he2016deep}
Kaiming He, Xiangyu Zhang, Shaoqing Ren, and Jian Sun,
\newblock ``Deep residual learning for image recognition,''
\newblock {\em CVPR}, pp. 770--778, 2016.

\bibitem{liu2016ssd}
Wei Liu, Dragomir Anguelov, Dumitru Erhan, Christian Szegedy, Scott Reed,
  Cheng-Yang Fu, and Alexander~C Berg,
\newblock ``Ssd: Single shot multibox detector,''
\newblock in {\em European conference on computer vision}. Springer, 2016, pp.
  21--37.

\bibitem{simonyan2014very}
Karen Simonyan and Andrew Zisserman,
\newblock ``Very deep convolutional networks for large-scale image
  recognition,''
\newblock {\em arXiv preprint arXiv:1409.1556}, 2014.

\bibitem{bethge2019back}
Joseph Bethge, Haojin Yang, Marvin Bornstein, and Christoph Meinel,
\newblock ``Back to simplicity: How to train accurate bnns from scratch?,''
\newblock {\em arXiv preprint arXiv:1906.08637}, 2019.

\bibitem{zhou2016dorefa}
Shuchang Zhou, Yuxin Wu, Zekun Ni, Xinyu Zhou, He~Wen, and Yuheng Zou,
\newblock ``Dorefa-net: Training low bitwidth convolutional neural networks
  with low bitwidth gradients,''
\newblock {\em arXiv preprint arXiv:1606.06160}, 2016.

\bibitem{wan2018tbn}
Diwen Wan, Fumin Shen, Li~Liu, Fan Zhu, Jie Qin, Ling Shao, and Heng Tao~Shen,
\newblock ``Tbn: Convolutional neural network with ternary inputs and binary
  weights,''
\newblock in {\em Proceedings of the European Conference on Computer Vision
  (ECCV)}, 2018, pp. 315--332.

\bibitem{zhang2019dabnn}
Jianhao Zhang, Yingwei Pan, Ting Yao, He~Zhao, and Tao Mei,
\newblock ``dabnn: A super fast inference framework for binary neural networks
  on arm devices,''
\newblock {\em arXiv preprint arXiv:1908.05858}, 2019.

\end{thebibliography}

\end{document}